\documentclass{ecai}

\usepackage{latexsym}
\usepackage{amssymb}
\usepackage{amsmath}
\usepackage{amsthm}
\usepackage{booktabs}
\usepackage{enumitem}
\usepackage{graphicx}
\usepackage{color}
\newcommand{\BibTeX}{B\kern-.05em{\sc i\kern-.025em b}\kern-.08em\TeX}
\usepackage{soul}

\usepackage{algorithm}
\usepackage{algpseudocode}
\usepackage{wrapfig}
\usepackage{caption,subcaption}
\usepackage{mathtools,amssymb,amsthm}
\usepackage{textcmds}
\usepackage{thm-restate}

\theoremstyle{definition}
\newtheorem{example}{Example}
\newtheorem{theorem}{Theorem}
\newtheorem{definition}{Definition}
\newtheorem{proposition}{Proposition}

\begin{document}

\begin{frontmatter}


\paperid{2315} 


\title{EXPLAIN, AGREE, LEARN: \\ Scaling Learning for Neural Probabilistic Logic}


\author[A]{\fnms{Victor}~\snm{Verreet}\thanks{Email: victor.verreet@kuleuven.be}}
\author[A]{\fnms{Lennert}~\snm{De Smet}}
\author[A]{\fnms{Luc}~\snm{De Raedt}}
\author[B]{\fnms{Emanuele}~\snm{Sansone}}

\address[A]{Department of Computer Science, KU Leuven, Leuven, Belgium}
\address[B]{Department of Electrical Engineering (ESAT), KU Leuven, Leuven, Belgium}

\begin{abstract}
Neural probabilistic logic systems follow the neuro-symbolic (NeSy) paradigm by combining the perceptive and learning capabilities of neural networks with the robustness of  probabilistic logic. Learning corresponds to likelihood optimization of the neural networks. However, to obtain the likelihood exactly, expensive probabilistic logic inference is required. To scale learning to more complex systems, we therefore propose to instead optimize a sampling based objective. We prove that the objective has a bounded error with respect to the likelihood, which vanishes when increasing the sample count. Furthermore, the error vanishes faster by exploiting a new concept of sample diversity. We then develop the EXPLAIN, AGREE, LEARN (EXAL) method that uses this objective. EXPLAIN samples explanations for the data. AGREE reweighs each explanation in concordance with the neural component. LEARN uses the reweighed explanations as a signal for learning. In contrast to previous NeSy methods, EXAL can scale to larger problem sizes while retaining theoretical guarantees on the error. Experimentally, our theoretical claims are verified and EXAL outperforms recent NeSy methods when scaling up the MNIST addition and Warcraft pathfinding problems.
\end{abstract}

\end{frontmatter}

\section{Introduction}

The field of neuro-symbolic (NeSy) artificial intelligence (AI) aims to combine the perceptive capabilities of neural networks with the reasoning capabilities of symbolic systems.
Many prominent NeSy systems achieve such a combination by attaching a probabilistic logic component to the neural output~\cite{manhaeve2018deep,yangneurasp}.
As a result, they tend to generalize better, can deal with uncertainty and require less training data compared to pure neural networks.
The main challenge that prevents the widespread adoption of NeSy is the difficulty to learn, because the learning signal for the neural network has to be propagated through the probabilistic logic component.

Existing work has tackled propagating the learning signal in a plethora of ways.
One line of work has tackled this challenge by relying on exact propagation using knowledge compilation.
This scales poorly for more complex systems because inference in probabilistic logic is \#P-complete~\cite{darwiche2002knowledge}.
Other lines of work propose different approximation schemes for the propagation.
A couple of approaches only consider the $k$-best solutions that satisfy the probabilistic logic component~\cite{manhaeve2021approximate,huang2021scallop}, but they introduce biases that can reinforce the (possibly incorrect) beliefs of the neural network.
More recently, A-NeSI~\cite{krieken2023nesi} proposed to use neural networks to provide a differentiable approximation of the logic component.
A-NeSI can provide some semantic guarantees, but it has to rely on extensive optimization and hyperparameter tuning to reduce the bias of its approximation.
While it still outperforms other methods in terms of scalability, the long training times of A-NeSI still limit its application to more complex problems.
None of the above methods provide bounds on the error of approximation.

To address these issues, this research focuses on scaling learning of the neural component in NeSy systems with a probabilistic logic component~\cite{manhaeve2018deep,winters2022stoch,krieken2023nesi} while providing strong statistical guarantees.
Concretely, we propose a surrogate objective to approximate the data likelihood, and prove it has some desirable properties that $k$-best or A-NeSI lacks.
First, it is an unbiased approximation of the likelihood.
Second, its approximation error is theoretically bounded.
Third, we show how this error can be decreased using a newly introduced concept of diversity.
Learning the neural component with this approximate objective sidesteps the requirement of differentiability for the probabilistic logic component that other methods focus on.

We also introduce the EXPLAIN, AGREE, LEARN (EXAL) method for constructing the surrogate objective. First, EXPLAIN samples explanations for the data at the level of the neural output, propagating the learning signal through the probabilistic logic component. We can control the resource scaling by choosing the number of samples. Then AGREE assigns an importance to each sampled explanation based on the predictions of the neural component. Together, EXPLAIN and AGREE construct the surrogate objective. LEARN then uses the surrogate objective to perform a classical learning iteration for the neural component with direct supervision on the neural output.

Experimentally, we validate our theoretical claims on synthetic data and show that all three steps in EXAL are necessary to achieve good performance. Moreover, we apply EXAL to two prominent NeSy problems, MNIST addition and Warcraft pathfinding and show that EXAL outperforms other NeSy methods such as A-NeSI~\cite{krieken2023nesi} in terms of execution time and accuracy for some instances when scaling to larger problem sizes.

\section{Background}

\begin{figure*}[t]
    \centering
    \includegraphics[scale=0.4]{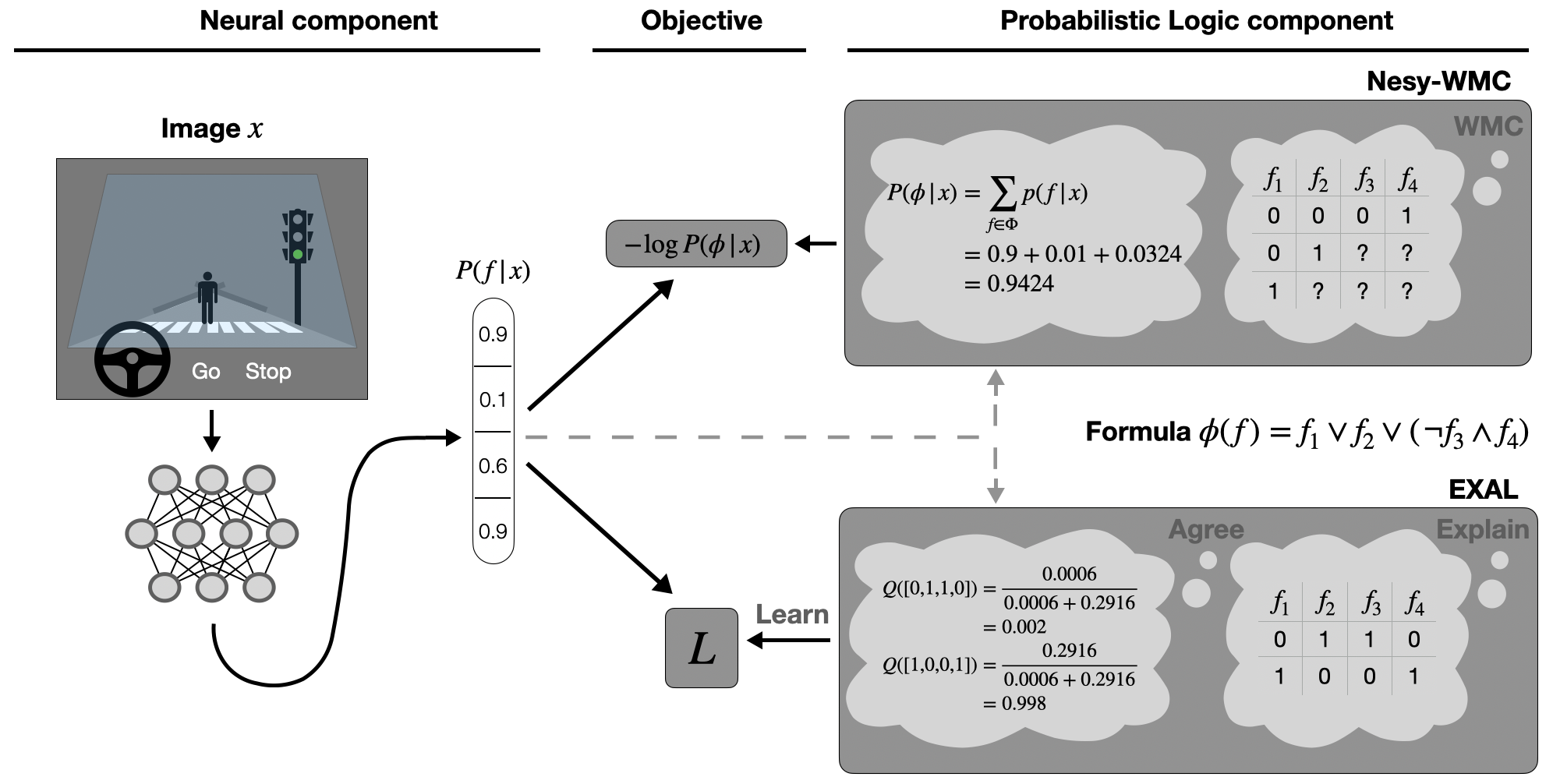}
    \caption{The neural component takes the input $x$ (left) to output probabilities over $n = 4$ variables. At the top, NeSy-WMC calculates the probability that $\phi$ is satisfied. Below, EXAL finds and reweighs explanations for $\phi$ to construct $L$.}
    \label{fig:car}
\end{figure*}

NeSy systems consist of both a neural and a symbolic component. In the case of neural probabilistic logic, the neural component handles perception of the world by mapping raw input $x \in \mathbb{R}^{d}$ to a probability distribution over $n$ binary variables $f \in \{ 0, 1 \}^{n}$. The symbolic component then performs logical reasoning over these binary variables to verify whether a formula $\phi$ is satisfied. More specifically, we choose weighted model counting (WMC) for the symbolic component, because probabilistic inference can be reduced to WMC~\cite{chavira2008}.

{\definition Let $\mathbb{A}(n) = \{ 0, 1, \mathord{?} \}^{n}$. We call $\mu \in \mathbb{A}(n)$ an \textit{assignment} of $n$ binary variables. If $\mu(k) = \mathord{?}$ for some $k$ then variable $k$ is not assigned a value and we call $\mu$ partial. Otherwise $\mu$ is complete.}

{\definition An \emph{explanation} of a formula $\phi$ is a complete assignment $\mu$ that satisfies $\phi$. Like~\cite{poole1993}, we use the term explanation, but it is also called a model. The set $\Phi$ contains all explanations of $\phi$.}

{\definition A WMC problem over $n$ binary variables $f \in \{ 0, 1 \}^{n}$ is defined by a propositional logical formula $\phi$, together with weight assignments $w_{k}: \{ 0, 1 \} \to \mathbb{R}^{+}$ for each variable.
The answer $W$ to the WMC problem is $W = \sum_{f \in \Phi} \prod_{k} w_{k}(f_{k})$~\footnote{The standard WMC formulation assumes independence of the variables $f_{k}$, but our theory still applies with a global weight $w(f)$ per explanation.}.
A NeSy-WMC problem is a WMC problem where the weights are computed by a neural network.} 

{\example \label{ex:car} The camera of a self-driving car provides an image with $d$ pixels as raw input. From this image, the neural component outputs $n = 4$ probabilities for detecting $f_{1}$ (pedestrian), $f_{2}$ (red light), $f_{3}$ (driving slow) and $f_{4}$ (crosswalk). The logical component decides the car should brake if there is a pedestrian or a red light or if it is driving too fast near a crosswalk. This is encoded in $\phi = f_{1} \lor f_{2} \lor (\lnot f_{3} \land f_{4})$. The weights of this NeSy-WMC problem are given by the neural component $w_k(f_k) = \text{n}_{k}(x)^{f_{k}} (1 - \text{n}_{k}(x))^{1 - f_{k}}$, where for example $\text{n}_{1}(x)$ is the probability that there is a pedestrian in the input image $x$ according to the neural network. The answer $W$ to the NeSy-WMC problem is the probability that the car should brake given the input image. This is illustrated in Fig.~\ref{fig:car}. We will use this as a running example throughout this paper.}

\section{The Objective}

\begin{figure}[t]
     \centering
     \includegraphics[width=.25\textwidth]{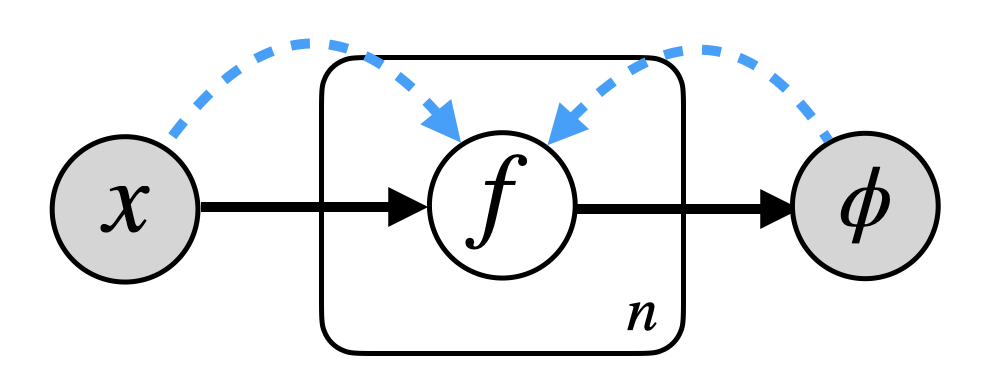}
     \caption{Probabilistic graphical model for NeSy-WMC. The full arrows follow the flow of information during inference. The dashed arrows follow the execution of EXAL, where explanations $f$ are sampled given $\phi$ and reweighed based on the neural output given $x$.}
     \label{fig:model}
     \vspace{1.5em}
\end{figure}

We now consider optimizing a NeSy-WMC model, consisting of a neural and logic component. We leave the neural component unspecified and only require it to be learnable via gradient descent. The data set contains $I$ tuples $\mathcal{D} = \{ (x_{i}, \phi_{i}) \mid i = 1, ..., I \}$, where $x_{i}$ is a raw input and $\phi_{i}$ is a logical formula that is assumed to hold for that data point and can be seen as its label. For example, if a self-driving car does not brake given image $x_{i}$ (see Ex.~\ref{ex:car}), then the label is $\phi_{i} = \lnot f_{1} \land \lnot f_{2} \land (f_{3} \lor \lnot f_{4})$, indicating there is no pedestrian, no red light and either there is no crosswalk or the car is driving slow enough. We will often drop the index $i$ of the data point.

Our setup is modelled by the probabilistic graphical model in Fig.~\ref{fig:model}.
It contains the input $x \in \mathbb{R}^{d}$, the $n$ binary variables $f \in \{ 0, 1 \}^{n}$ and whether $\phi$ is satisfied.
The logic component is deterministic $P(\phi \mid f) = 1$ if $f$ satisfies $\phi$ and zero otherwise.
The neural component outputs a distribution $P(f \mid x)$, which are the weights of the NeSy-WMC problem.
The answer $W$ of the NeSy-WMC problem is the probability $P(\phi \mid x)$.
The neural component is usually learned via likelihood maximization on the data set $\mathcal{D}$.
However, calculating the likelihood requires expensive inference, which is intractable for complex systems.
We therefore propose the following surrogate objective $L$ that bounds the true likelihood.

{\definition \label{def:objective} For a data point $(x_{i}, \phi_{i})$ let $Q_{i}$ be an auxiliary distribution over the explanations $\Phi_{i}$ of the label $\phi_{i}$ and let $P_{i}$ be the distribution over the explanations that the neural component outputs given $x_{i}$. We then define the surrogate objective $L = \sum_{i} \text{KL}(Q_{i} \mid P_{i})$.}

Intuitively, the auxiliary distributions $Q_{i}$ propose explanations for the label $\phi_{i}$, from which the neural component $P_{i}$ can learn (Fig.~\ref{fig:model}). We now show the bound on the negative log-likelihood loss. The full derivation of this bound can be found in the appendix.
\begin{align}
    -\log P(\mathcal{D}) & \leq -\sum_{i} \sum_{f \in \Phi_{i}} Q_{i}(f) \log \frac{P(f \mid x_{i})}{Q_{i}(f)} \nonumber \\
    & = \sum_{i} \text{KL}(Q_{i} | P_{i}) = L \quad . \label{eq:bound}
\end{align}
The tightest bound in Eq.~\ref{eq:bound} is obtained by minimizing the Kullback-Leibler (KL) divergence with respect to both $P_{i}$ and $Q_{i}$. The learning signal for the neural component $P_{i}$ is thus provided by $Q_{i}$, sidestepping the requirement of backpropagating gradients through the logic component~\cite{manhaeve2018deep,xusemantic,ahmed2022semantic}. Moreover, we also avoid expensive exact inference based on knowledge compilation~\citep{oztok-kc} when learning with $L$.

\section{EXPLAIN, AGREE, LEARN}

The EXPLAIN, AGREE, LEARN method aims to minimize the objective $L$ from Def.~\ref{def:objective}.
This objective contains a KL divergence between the distributions $P_i$ and $Q_i$, raising a number of issues.
The first issue is that each $Q_i$ is supported on all explanations of the formula $\phi_i$, which are unknown a priori and infeasible to obtain for complex problems.
Hence, we introduce the EXPLAIN algorithm that \emph{samples} explanations of $\phi_i$ used to approximate the support of $Q_i$.
After sampling explanations, the samples are reweighted by the neural component $P_i$ in the AGREE step such that samples deemed more likely by $P_i$ have a higher weight.
Alternatively, the AGREE step can be interpreted as a minimization of $L$ with respect to all $Q_i$.
The EXPLAIN and AGREE steps both serve to construct a suitable proposal distribution $Q_i$ for the objective $L$.
$L$ is then used to optimize each $P_i$ in the LEARN step, which performs a traditional update step using backpropagation.

\subsection{EXPLAIN: Sampling Explanations}

The first step to minimizing $L$ is to construct the support of the proposal distributions $Q$.
As the support of each $Q$ consists of explanations of the logical formula $\phi$, we propose the EXPLAIN algorithm to sample such explanations.
EXPLAIN is a stochastic variant of the Davis–Putnam–Logemann–Loveland (DPLL) algorithm, for which we first introduce some necessary concepts.

{\definition A \textit{sampling strategy} $\Sigma(\mu)$ for an assignment $\mu$ is a probability distribution over the set of all assignments obtained from $\mu$ by assigning one more variable a value.}

EXPLAIN is shown in Alg.~\ref{alg:explain} and differs from traditional DPLL on lines \ref{alg:explain-sample} and \ref{alg:explain-recurse}. Specifically, instead of iterating over all possible assignments, a new assignment $\mu'$ is sampled according to $\Sigma(\mu)$. EXPLAIN starts with the formula $\phi$, assumed to be in conjunctive normal form, the \emph{empty assignment} $\mu_{?}$, defined as $\forall k : \mu_{?}(k) = \mathord{?}$, and a uniform sampling strategy $\Sigma(\mu)$. First, trivial variable assignments, e.g. unit clauses, are propagated to $\mu$. The algorithm then terminates if $\phi$ is satisfied or restarts if $\phi$ is unsatisfiable. Otherwise, a value for one more variable is sampled using $\Sigma(\mu)$ and EXPLAIN is recursively called. Notice how the distribution of returned explanations is entirely determined by the sampling strategy $\Sigma(\mu)$.

\example{Recall Ex.~\ref{ex:car} and Fig.~\ref{fig:car}. If the self-driving car brakes at some point, the corresponding query is $\phi(f) = f_{1} \lor f_{2} \lor (\lnot f_{3} \land f_{4})$. EXPLAIN can sample multiple explanations for this query, e.g. $\mu_{1} = (0, 1, 1, 0)$ (the car brakes due to a red light) or $\mu_{2} = (1, 0, 0, 1)$ (the car brakes due to a pedestrian on a crosswalk).}

\begin{algorithm}[t]
\caption{Sampling algorithm $\text{EXPLAIN}(\phi, \mu, \Sigma)$}
\label{alg:explain}
\begin{algorithmic}[1]
    \State \textbf{Input:} formula $\phi$, assignment $\mu$, sampling strategy $\Sigma$
    \State \textbf{Output:} explanation for $\phi$
    \State
    \State $\mu \leftarrow \text{propagate}(\phi, \mu)$ \Comment{trivial assignments}
    \If{$\mu \text{ is an explanation for } \phi$}
        \State \textbf{return} $\mu$ \Comment{explanation found}
    \ElsIf{$\mu \text{ cannot satisfy } \phi$}
        \State \textbf{return} $\text{EXPLAIN}(\phi, \mu_{?}, \Sigma)$ \Comment{restart due to conflict}
    \EndIf
    \State $\mu' \leftarrow \text{sample from } \Sigma(\mu)$ \label{alg:explain-sample} \Comment{new variable assignment}
    \State \textbf{return} $\text{EXPLAIN}(\phi, \mu', \Sigma)$ \label{alg:explain-recurse} \Comment{recurse}
\end{algorithmic}
\end{algorithm}

During the execution of EXPLAIN it is possible to encounter conflicting variable assignments.
Specifically, a conflict occurs when the query $\phi$ is unsatisfiable by $\mu$. The execution of EXPLAIN is stochastic because of the sampling strategy $\Sigma$, so it is possible that some sampled assignments lead to a conflict whereas others do not. There are several policies for dealing with conflicts. The easiest policy is to simply stop execution and restart EXPLAIN from the top. Another policy is to backtrack and sample a different new variable assignment $\mu'$. Furthermore, a limit on the amount of backtracking can be imposed before deciding to restart. Although Alg.~\ref{alg:explain} restarts execution upon encountering a conflict, our implementation employs backtracking to the latest sampled assignment.
Note that, if backtracking and a deterministic sampling strategy are used, EXPLAIN reduces to DPLL.

\subsection{AGREE: Updating the Weights}

We can choose any auxiliary distribution $Q$ over explanations of $\phi$ in Def.~\ref{def:objective} of $L$ and will choose $Q$ to minimize $L$.
The explanations sampled by the EXPLAIN algorithm define the support of $Q$ and varying $Q$ amounts to reweighing every sample.
It turns out that the objective $L$ is minimized with respect to these weights $Q$ when they are proportional to the output of the neural component $P(f \mid x)$. Intuitively, explanations that the neural component deems more likely, should be assigned a higher weight to minimize $L$.
We now formalize this result and provide a bound on the error of the approximation.

{\theorem \label{th:estimator} 
Let $\Psi \subseteq \Phi$ be the explanations sampled by EXPLAIN and define $Q^\ast(f) = P(f \mid x) / \sum_{f' \in \Psi} P(f' \mid x)$, then:
\begin{enumerate}
    \item \textbf{Optimality.} $Q^\ast$ globally minimizes $L$ with respect to $Q$.
    \item \textbf{Optimum.} The minimum of $L$ is given by $-\sum_{i} \log P^\ast(\phi_{i} \mid x_{i})$, with $P^\ast(\phi_{i} \mid x_{i}) = \sum_{f \in \Psi_{i}} P(f \mid x_{i})$.
    \item \textbf{Bounds.} For any $x$ and $\phi$ the true probability is bounded by $P^\ast(\phi \mid x) \leq P(\phi \mid x) \leq 1 - P^\ast(\lnot \phi \mid x)$.
\end{enumerate}

\begin{proof}
\textbf{Optimality.} Let us write out one term of $L$, dropping the data point index $i$, yielding
\begin{equation}
    \text{KL}(Q \mid P) = -\sum_{f \in \Psi} Q(f) \log \frac{P(f \mid x)}{Q(f)} \quad . \label{eq:kl}
\end{equation}
Next, we use the method of Lagrange multipliers to impose the constraint $\sum_{f \in \Psi} Q(f) = 1$ while minimizing this KL divergence with respect to all the $Q(f)$.
This method leads to the equations
\begin{align}
    & \frac{\partial}{\partial Q(f)} \left[ -\sum_{f' \in \Psi} Q(f') \log \frac{P(f' \mid x)}{Q(f')} + \lambda \left( \sum_{f' \in \Psi} Q(f') - 1 \right) \right] \nonumber \\
    & = -\log \frac{P(f \mid x)}{Q(f)} + \lambda + 1 = 0,
\end{align}
and thus $Q(f) = \exp(\lambda + 1) P(f \mid x)$.
From the constraint we solve $\lambda = -1 - \log \sum_{f \in \Psi} P(f \mid x)$ and filling this into the expression for $Q(f)$ gives the optimum at $Q^\ast(f) = P(f \mid x) / \sum_{f' \in \Psi} P(f' \mid x)$.

\textbf{Optimum.} Filling in the optimal value $Q^\ast(f)$ in Eq.~\eqref{eq:kl} gives
\begin{equation}
    \text{KL}(Q \mid P) = -\log \sum_{f \in \Psi} P(f \mid x) = -\log P^\ast(\phi \mid x)
\end{equation}
and summing over the data index $i$ gives the result.

\textbf{Bounds.} Since $\Psi \subseteq \Phi$, the inequality
\begin{equation}
    P^\ast(\phi \mid x) = \sum_{f \in \Psi} P(f \mid x) \leq \sum_{f \in \Phi} P(f \mid x) = P(\phi \mid x)
\end{equation}
holds. The upper bound on $P(\phi \mid x)$ is the same inequality applied to $\lnot \phi$ and using $P(\phi \mid x) = 1 - P(\lnot \phi \mid x)$.
\end{proof}}

The first result tells us that we can solve the minimization of $L$ with respect to $Q$ analytically, avoiding iterative optimization methods. The value reached by the global minimizer is the estimator of the true log-likelihood that we use during the LEARN step. The third result bounds the error of the approximate objective with respect to the true log-likelihood. These bounds get tighter with more samples and eventually become exact when all possible explanations of the query have been sampled at least once, i.e., whenever $\Psi = \Phi$. This convergence of bounds is also shown experimentally in Sec.~\ref{sec:exp-bounds}.

{\example Assume we have sampled two explanations $\mu_{1}$ (the car brakes due to a red light) and $\mu_{2}$ (the car brakes due to a pedestrian on a crosswalk). The support of $Q$ is then $\Psi = \{ \mu_{1}, \mu_{2} \}$ with weights $Q(\mu_{1})$ and $Q(\mu_{2})$ summing to unity. The AGREE step sets the weights in accordance to how likely the neural component $P$ deems each explanation, i.e. $Q(\mu_{1}) = P(\mu_{1} \mid x) / ( P(\mu_{1} \mid x) + P(\mu_{2} \mid x) )$, with $P(\mu_{1} \mid x)$ the probability that there is a red light in image $x$. See also Fig.~\ref{fig:car}.}

\subsection{LEARN: Updating the Neural Component}

The LEARN step minimizes $L$ with respect to the parameters of the neural component $P$ using standard gradient descent techniques. When $|\Psi| = 1$, we observe that the loss $L$ is equivalent to the traditional cross-entropy loss. We can interpret the LEARN step as a standard supervised learning step, where the labels are provided by the EXPLAIN and AGREE steps. The full EXAL algorithm is shown in Alg.~\ref{alg:exal}.

\begin{algorithm}[t]
    \caption{NeSy-WMC learning method $\text{EXAL}(p, \mathcal{D})$}
    \label{alg:exal}
\begin{algorithmic}[1]
    \State \textbf{Input:} neural parameters $p$, dataset $\mathcal{D}$
    \State\textbf{Output:} updated $p$
    \State $(x, \phi) \leftarrow \text{data point from } \mathcal{D}$
    \State $\Psi \leftarrow \{ \}$
    \For{$t = 1, ..., T$}
        \State $\Psi \leftarrow \Psi \cup \{ \text{EXPLAIN}(\phi, \mu_{?}, \Sigma) \}$ \Comment{EXPLAIN}
    \EndFor
    \State $L \leftarrow \log \sum_{f \in \Psi} P(f \mid x, p)$ \Comment{AGREE}
    \State $p \leftarrow p + \eta \nabla_{p} L$ \Comment{LEARN}
    \State \textbf{return} $p$
\end{algorithmic}
\end{algorithm}

\section{Optimizations using Diversity}

The following section provides an optimization for EXAL using the concept of diversity. The formula $\phi$ of a data point can have many explanations and the true underlying explanation is not known.
Consequently, sampling a diverse set of explanations with EXPLAIN can increase the chance of seeing the true explanation. For example, a self-driving car will more often brake because for a pedestrian than an animal on the road. An animal should however not be dismissed as a possible explanation for braking and could still be the true explanation for a certain data point.
We formalise this intuition through the notion of diversity of explanations.
Additionally, we provide two practical methods to increase the diversity of the samples obtained from the EXPLAIN algorithm.

\subsection{Diversity of Explanations} \label{sec:diversity}

We formally define diversity and prove in Prop.~\ref{prop:diversity} that sampling a more diverse set of explanations leads to a tighter bound $L$ on the negative log-likelihood (see Eq.~\ref{eq:bound}).

{\definition The \emph{diversity} of a set $\Psi$ of assignments with respect to $\phi$ is $\delta(\Psi) = |\Psi \cap \Phi|$, where $\Phi$ are the explanations of $\phi$.}

{\proposition \label{prop:diversity} If $\Psi_{1} \subseteq \Psi_{2}$ then $\delta(\Psi_{1}) \leq \delta(\Psi_{2})$ and $L_{2} \leq L_{1}$ where $L_{1}$ and $L_{2}$ are the objectives constructed from Equation~\eqref{eq:kl} using $\Psi_{1}$ and $\Psi_{2}$, respectively (see optimum in Thm.~\ref{th:estimator}).
\begin{proof}
If $\Psi_{1} \subseteq \Psi_{2}$ then $\sum_{f \in \Psi_{1}} P(f \mid x) \leq \sum_{f \in \Psi_{2}} P(f \mid x)$. Taking the negative logarithm of both sides gives the result
\begin{equation}
    L_{2} = -\log \sum_{f \in \Psi_{2}} P(f \mid x) \leq -\log \sum_{f \in \Psi_{1}} P(f \mid x) = L_{1} \quad .
\end{equation}
\end{proof}}

The following paragraph explains how to increase diversity with a suitable sampling strategy. Recall that the execution of EXPLAIN depends entirely on the sampling strategy $\Sigma(\mu)$. We parametrize $\Sigma_{\theta}(\mu)$ with parameters $\theta \in \mathbb{R}^{+}$ and change $\theta$ to increase the diversity. Concretely, we keep track of how often $\Sigma_{\theta}(\mu)$ samples a (partial) assignment $\mu'$ during the execution of EXPLAIN. Call this count $N(\mu')$. Every time $\mu'$ is sampled, the probability to sample $\mu'$ will be reduced by a factor of $\exp(-\theta)$, encouraging the exploration of different assignments.
In other words, the probability that $\Sigma_{\theta}(\mu)$ samples $\mu'$ is chosen proportional to $\exp(-N(\mu') \theta)$.
Uniform sampling without diversity optimizations corresponds to $\theta = 0$.

\subsection{Reformulation as a Markov Decision Process} \label{sec:mdp}

A recent result \cite{bengio2021flow} of generative flow networks (GFlowNets) for Markov decision processes (MDPs) also allows us to increase the diversity of sampled explanations through optimisation of $\theta$.
Their results include a method to maximize the expected reward when executing an MDP with certain properties. If we set the reward in the MDP equal to the diversity, their method can be used to increase the diversity. To apply the result, running EXPLAIN $T$ times for a query $\phi$ to obtain a set of $T$ samples first has to be framed as an MDP. Recall that $\mathbb{A}(n) = \{ 0, 1, ? \}^{n}$ is the set of all assignments to $n$ binary variables.
Therefore, the MDP has states $(\mu, t, \Psi) \in \mathbb{A}(n) \times \mathbb{N} \times 2^{\mathbb{A}(n)}$, where $\mu$ is the current assignment in the $t^{\text{th}}$ run of EXPLAIN and $\Psi$ contains the assignments returned by previous runs of EXPLAIN. The initial state is $(\mu_{?}, 0, \{ \})$. For an intermediate state $(\mu, t, \Psi)$ with partial $\mu$, the possible actions are to assign a value to an unassigned variable of $\mu$, leading to a new state $(\mu', t, \Psi)$. If $\mu$ is complete, the only action is to go to the next run of EXPLAIN, transitioning to the state $(\mu_{?}, t + 1, \Psi \cup \{ \mu \})$. The terminal states are $(\mu_{?}, T, \Psi)$ which have a reward $R(\mu_{?}, T, \Psi) = \delta(\Psi)$. All other states have no reward.

{\proposition \label{prop:termination} The transition graph of the above MDP is acyclic and the MDP will always terminate.

\begin{proof}
For a state $(\mu, t, \Psi)$ we define a quantity, which we call the \textit{progress}, as $(n + 1) t + |\{ k \mid \mu(k) \neq \mathord{?} \}|$ with $n$ the number of variables. We show that the progress increases by a value of one for every transition in the MDP. There are two cases. If $\mu$ is partial, the only action is to assign a value to a variable. This leaves the first term of the progress unchanged, but increases the second term by one. If $\mu$ is complete, the MDP transitions to $(\mu_{?}, t + 1, \Psi)$. The second term changes by $-n$, but the increment of $t$ causes the first term to increase by $n + 1$, again resulting in a net gain of one. Since the progress always increases when transitioning from state to state, no previous state can ever be visited. Hence the MDP is acyclic. Furthermore all states with progress $(n + 1) T$ are terminal states, so the MDP is guaranteed to terminate.
\end{proof}}

The execution of EXPLAIN, governed by the sampling strategy $\Sigma(\mu)$, translates to a probabilistic policy for the MDP as follows. In a state $(\mu, t, \Psi)$ with partial $\mu$, the probability to take the action transitioning to $(\mu', t, \Psi)$ is chosen equal to the probability of $\mu'$ as given by $\Sigma(\mu)$. For complete $\mu$, the only possible action is to rerun EXPLAIN. There is thus a mapping from sampling strategies $\Sigma(\mu)$ to MDP policies.
This mapping allows the maximization of the diversity of parametrized sampling strategies $\Sigma_{\theta}(\mu)$ via GFlowNets.

{\theorem Consider the flow function $F: (\mathbb{A}(n) \times \mathbb{N} \times 2^{\mathbb{A}(n)})^{2} \to \mathbb{R}^{+}$ that solves the flow equation
\begin{equation}
    \sum_{s' \to s} F(s', s) = R(s) + \sum_{s \to s'} F(s, s') \label{eq:flow}
\end{equation}
where the sums run over all states $s' = (\mu', t', \Psi')$ incoming into $s = (\mu, t, \Psi)$ or outgoing from $s$ respectively. Following the policy that transitions from $s$ to $s'$ with probabilities proportional to $F(s, s')$, the probability to terminate in $s$ is proportional to $R(s) = R(\mu, t, \Psi) = \delta(\Psi)$, i.e. the diversity of $\Psi$.

\begin{proof}
Because the MDP is acyclic and guaranteed to terminate (Prop.~\ref{prop:termination}), this result follows by applying Prop.~2 in \cite{bengio2021flow} to the MDP.
\end{proof}}

The above theorem provides a way to optimize the expected diversity when running EXPLAIN, namely by choosing $\theta$ such that the policy given by $\Sigma_{\theta}(\mu)$ is as close as possible to the policy given by $F$. The solution $F$ to the flow equation can be difficult to find, but \citet{bengio2021flow} provide one method. The general idea is to minimize the logarithmic difference $\Delta$ between the left- and right-hand side of the flow equation (Eq.~\ref{eq:flow})
\begin{equation}
    \Delta(F) = \sum_{i} \left[ \log \left( \frac{\sum_{s \to s_{i}} F(s, s_{i})}{R(s_{i}) + \sum_{s_{i} \to s} F(s_{i}, s)} \right) \right]^{2}
\end{equation}
with the sum running over states $s_{i}$ in trajectories sampled from the MDP. We experimentally compare this optimization approach to the simpler approach at the end of Sec.~\ref{sec:diversity}. The results are discussed in Sec.~\ref{sec:exp-diversity}.

\section{Experiments}

Experiments are provided to support our theoretical claims. We show that we can learn to generate diverse samples using the EXPLAIN algorithm. Then the importance of diversity and the AGREE step are illustrated by looking at the convergence of the learning objective. Lastly, we apply the EXAL method to the MNIST addition and Warcraft pathfinding tasks. Our implementation can be found here  ( \url{https://anonymous.4open.science/r/exal-526C} ) and more details about the experiments are in the appendix.

\subsection{Diversity (EXPLAIN)} \label{sec:exp-diversity}

For the first experiment, we want to investigate how different sampling strategies compare in terms of diversity. To this aim, we run EXPLAIN on formulas generated in 3 different classes and with a varied number of variables, ranging between 15 and 60. Generation details for each class $branch$, $split$ and $bottom up$ are in the appendix. For every generated formula, we evaluate the diversity of different sampling strategies so they can be directly compared. This experiment is concerned only with logical explanations and does not use neural networks. We count how many unique explanations are sampled during subsequent calls to EXPLAIN, i.e. the diversity. The counts at each time step are averaged out over 200 runs on the same formula. This is done for fixed values of $\theta$, as well as for parameters $\theta$ that are learned using the diversity algorithm in Sec.~\ref{sec:mdp}. As baselines, a uniform sampler and the theoretical maximally diverse sampler are considered. 

\begin{figure*}
    \centering
    \includegraphics[width=0.9\linewidth]{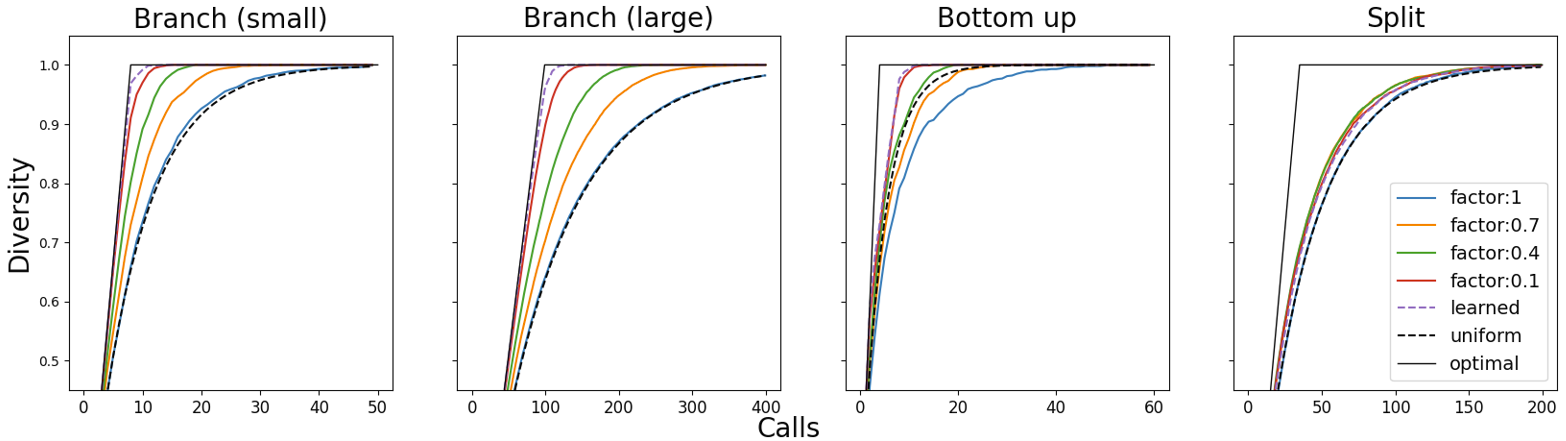}
    \caption{Diversity as a function of EXPLAIN calls for different formulas and sampling strategies. The factor is $\exp(-\theta) \in \{ 0.1, 0.4, 0.7, 1 \}$.}
    \label{fig:diversity}
\end{figure*}

Fig.~\ref{fig:diversity} shows the evolution of the diversity for each strategy. The diversity has been normalized to the unit interval for easy comparison. Although the EXPLAIN algorithm is not guaranteed to sample uniformly, in many instances it is close to uniform. This can be seen by the overlapping of the uniform line and the line with $\theta = 0$. For the $bottom up$ formulas however, EXPLAIN is not uniform and performs worse for $\theta = 0$. The diversity of EXPLAIN is typically increased by choosing a larger $\theta$ and can get close to the optimal diversity. In some instances, such as in the $split$ formulas, having $\theta > 0$ increases diversity, but there is little difference in diversity for the different values of $\theta$. We also see that learning to optimize the diversity performs better than sampling uniformly. It is recommended to choose a higher $\theta$ for improving the diversity, but stability issues can occur for too high values of $\theta$.

\subsection{Convergence of Bounds (AGREE)} \label{sec:exp-bounds}

\begin{figure}[ht]
    \centering
    \begin{subfigure}{0.45\textwidth}
        \centering
        \includegraphics[width=\linewidth]{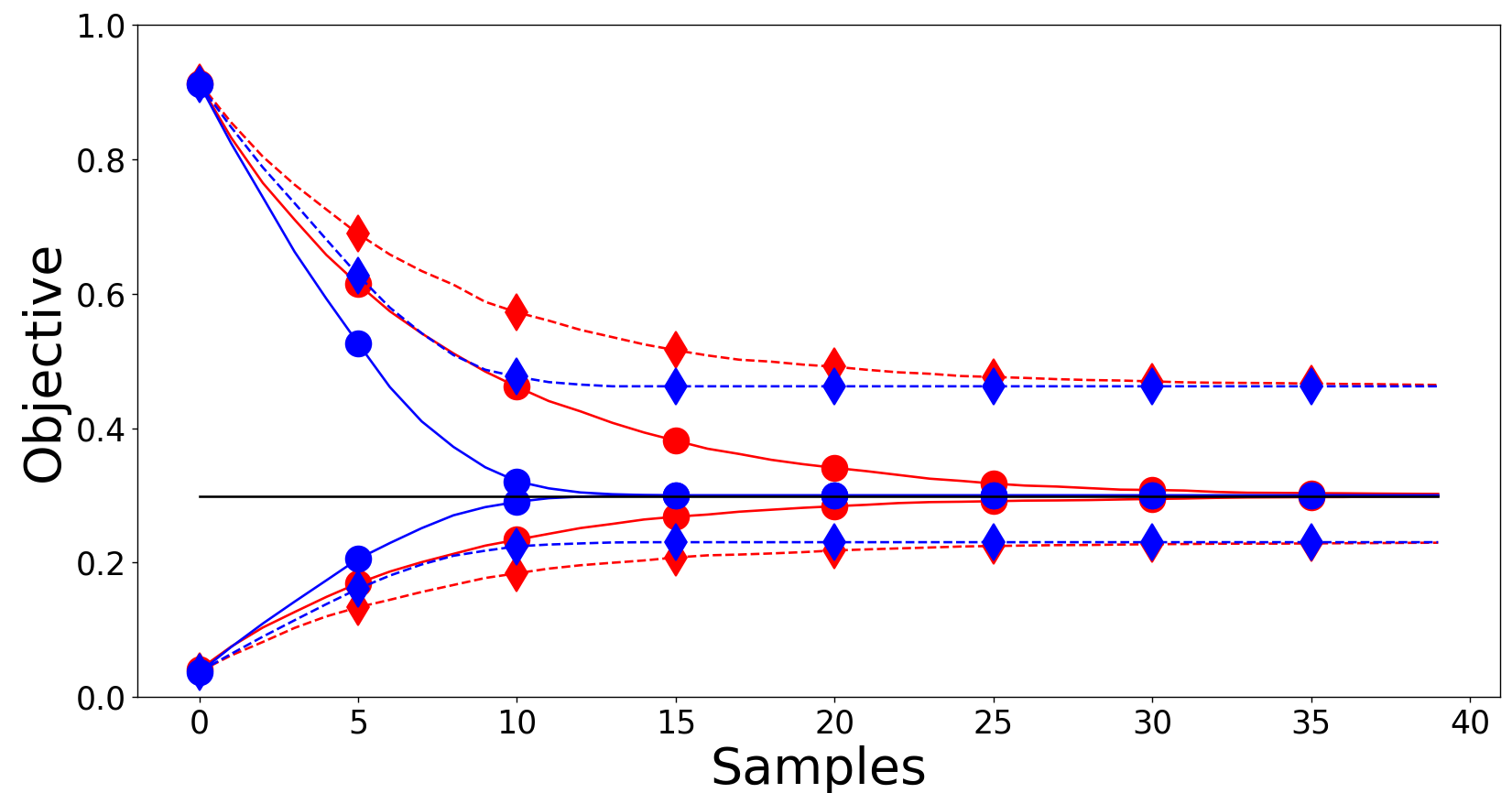}
    \end{subfigure}
    \begin{subfigure}{0.45\textwidth}
        \centering
        \includegraphics[width=\linewidth]{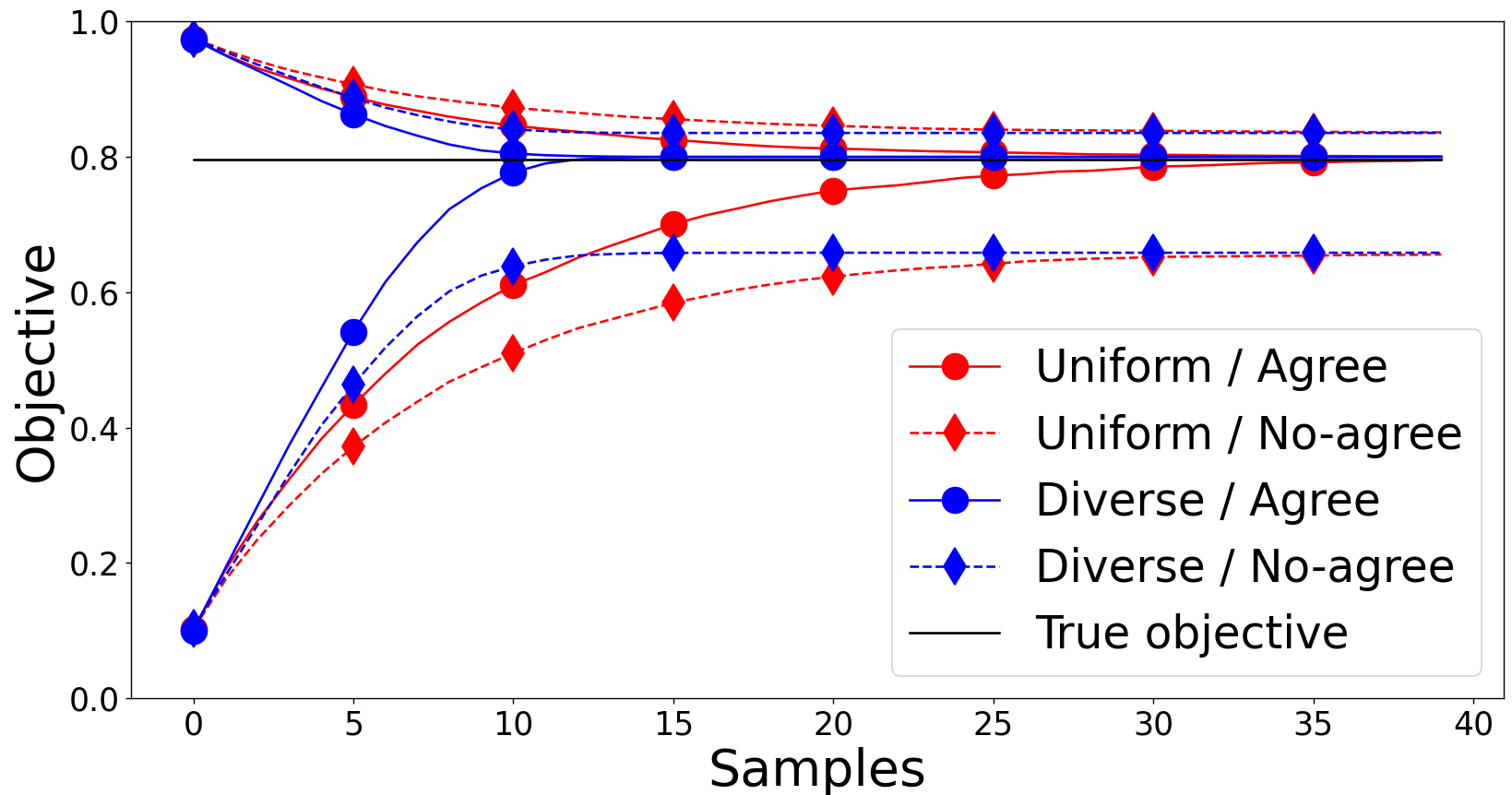}
    \end{subfigure}
    \caption{Convergence of objectives for increasing sample count.}
    \label{fig:bounds}
    \vspace{1em}
\end{figure}

We now experimentally answer what the importance of diversity and of the AGREE step are. For this purpose we compare the surrogate objective with the true likelihood for different number of samples. More concretely, explanations are sampled for the formula and for its negation, using both the diverse and non-diverse versions of EXPLAIN. Then the objectives with and without the AGREE step are calculated. The AGREE step makes use of the probabilities given by the neural network. We perform this synthetic experiment without neural networks by generating the probabilities uniformly at random on the unit interval. The results for all four combinations are shown in Figure~\ref{fig:bounds}. The samples for the formula give a lower bound, whereas samples for its negation give an upper bound. The AGREE step is necessary for the objectives to converge to the true likelihood. Furthermore, convergence happens faster when the samples are more diverse, confirming our theoretical results that diversity improves the error bounds.

\subsection{MNIST Addition (EXAL)}

We evaluate our approach on the standard NeSy experiment of learning to sum MNIST digits~\citep{manhaeve2018deep,manhaeve2021approximate}. The input to this task is two sequences of $N$ MNIST images, where each sequence represents a decimal number of length $N$. The desired output is the sum of these two numbers, which is also the only supervision. The sum is encoded in a logical formula. For example, if two single-digit numbers $a$ and $b$ add up to 3, the formula is $\phi = (a=0 \land b=3) \lor (a=1 \land b=2) \lor (a=2 \land b=1) \lor (a=3 \land b=0)$. The size of the possible assignments of values to each digit in each sequence ($10^{2N}$) in combination with the distant supervision is what makes learning to sum MNIST digits a challenging task. Although our general implementation of EXPLAIN is not parallelized, we have used a parallelized implementation of EXPLAIN that can sample digits for this experiment. More details on the experimental setup and hyperparameters is given in the appendix.

\begin{table}[ht]
    \caption{Test accuracy of predicting the correct sum of two sequences of MNIST digits of length $N$. Accuracies of A-NeSI, DeepStochLog and Embed2Sym were taken as reported by~\cite{krieken2023nesi}. Additionally, we report the time in terms of minutes for EXAL and the main competitor, A-NeSI, using the exact same computational resources.}
    \label{tab:mnist}
    \centering
    \resizebox{\linewidth}{!}{
    \begin{tabular}{l|ccc}
    \toprule
    \textbf{Method} & $N = 2$ & $N = 4$ & $N = 15$ \\
    \midrule
    Reference & $96.06$ & $92.27$ & $73.97$ \\
    \midrule
    DeepStochLog & $96.40 \pm 0.10$ & $ 92.70 \pm 0.60$ & T/O \\
    Embed2Sym & $93.81 \pm 1.37$ & $91.65 \pm 0.57$ & $60.46 \pm 20.4$ \\
    A-NeSI & $95.96 \pm 0.38$ & $92.56 \pm 0.79$ & $75.90 \pm 2.21$ \\
    \textbf{EXAL} & $95.82 \pm 0.36$ & $91.77 \pm 0.83$ & $73.27 \pm 2.05$ \\
    \midrule
    A-NeSI (time) & $81.7$ & $198.2$ & $1979.9$ \\
    \textbf{EXAL} (time) & $51.4 \pm 2.3$ & $74.8 \pm 7.2$ & $198.6 \pm 15.7$ \\
    \bottomrule
    \end{tabular}}
\end{table}

Table~\ref{tab:mnist} shows how we compare to the state of the art for $N = 2$, $N = 4$ and $N = 15$ digits. EXAL is competitive with A-NeSI \cite{krieken2023nesi} and provides state-of-the-art performance, with the desired reference accuracy always within margin of error. This reference expresses the expected performance of predicting the correct sum of two $N$ digit numbers given a $99\%$ accurate digit classifier. Additionally, EXAL does not require fine-tuning of a neural approximation of the logic component or defining appropriate priors up front, in contrast to A-NeSI. Instead, it exploits explanations to directly acquire a suitable proposal distribution for learning.

This suitable proposal translates into an overall procedure that is scalable and sample efficient. Looking at the reported average runtimes in Table~\ref{tab:mnist} for EXAL and A-NeSI, directly leveraging explanations instead of relying on approximation through optimisation can result in an order of magnitude faster learning times. Interestingly, the difference in learning time becomes larger as the problem size increases.

\subsection{Warcraft Pathfinding (EXAL)}

The task of Warcraft pathfinding, as described in \cite{vlastelica}, is to find the shortest path between two corner points of a two-dimensional grid given an image representing the grid. The exact traversal costs of each node are not known and have to be predicted from the image by the neural component. The only supervision for training the neural component is the true shortest path in the grid.

In this context, the symbolic component is a shortest path finding algorithm, e.g. Dijkstra's algorithm, and an explanation is a cost assignment to every node in the grid such that the shortest path given by those costs coincides with the true shortest path. Sampling explanations is done as follows. First the grid is initialized with all nodes set to the highest cost, except for nodes on the true shortest path, which are set to the lowest cost. This guarantees that the given shortest path coincides with the true shortest path. Then a Gibbs sampling procedure is executed that resamples node costs with the restriction that the shortest path is unchanged. After a burn-in period of $100$ resampling steps, the resampled node costs are used as a training signal for the neural component.

Once the neural component has been trained to predict node costs, the NeSy system can predict the shortest path given an image of a grid. A prediction is only considered correct if the predicted shortest path overlaps entirely with the true shortest path.

\begin{table}[t]
    \caption{Test accuracy of predicting the shortest path given an image of a grid. Results of RLOO, A-NeSI and A-NeSI with RLOO were taken as reported by~\citet{krieken2023nesi}. Runtimes are reported in minutes. It should be noted that the runtimes from A-NeSI and EXAL were measured on different machines. EXAL has been executed on a Dell XPS 15 (i7, 16GB, 512GB, FHD, GPU) laptop.}
    \label{tab:warcraft}
    \centering
    \resizebox{0.8\linewidth}{!}{\begin{tabular}{l|cc}
    \toprule
    \textbf{Method} & $12 \times 12$ & $30 \times 30$\\
    \midrule
    RLOO & $43.75 \pm 12.35$ & $12.59 \pm 16.38$ \\
    A-NeSI & $94.57 \pm \phantom{0}2.27$ & $17.13 \pm 16.32$ \\
    A-NeSI + RLOO & $98.96 \pm \phantom{0}1.33$ & $67.57 \pm 36.76$ \\ 
    \textbf{EXAL} & $94.19 \pm \phantom{0}1.74$ & $80.85 \pm \phantom{0}3.83$ \\
    \midrule
    A-NeSI (time) & $1380$ & $2640$ \\
    \textbf{EXAL} (time) & $11.1 \pm \phantom{0}0.1$ & $84.3 \pm \phantom{0}0.7$ \\
    \bottomrule
    \end{tabular}}
\end{table}

Two main conclusions can be drawn from the reported test accuracies in Table~\ref{tab:warcraft}. First, the same trend from the MNIST experiment continues, namely that EXAL provides orders of magnitude faster learning times compared to the state-of-the-art. Second, this quick convergence time does not impact the acquired accuracies. Even more, for the most challenging $30 \times 30$ grid, EXAL significantly outperforms A-NeSI. Together, they provide strong evidence that using explanations improves scaling of NeSy learning.

\section{Related Work}
We review recent works on neurosymbolic (NeSy) learning and scalable logical inference. For a more detailed discussion on NeSy, we refer the reader to recent surveys in~\cite{besold2021neuro,giunchiglia2022deep}.

\textbf{NeSy relaxations}. Several NeSY systems address the scalability of inference with continuous relaxations based on fuzzy logic semantics~\citep{badreddine2022logic,giunchiglia2021multi,daniele2019knowledge,li2019augmenting,gan2021judgment,sachan2018learning,donadello2019compensating}. However, these relaxations may introduce approximations that can yield different outputs for equivalent logic formulas~\citep{krieken2022analyzing,grespan2021evaluating}. 
 In contrast, our work performs inference and learning without resorting to such relaxations while retaining a probabilistic interpretation. Similar to us, the work in~\cite{wang2023grounding} provides a NeSy solution based on SMT solvers, which does not require backpropagation through the symbolic component. The solution approximates the gradient by discretizing the neural representation and solving a combinatorial optimization problem. In contrast, our work does not require to approximate the gradient as training proceeds in a supervised fashion thanks to the sampled explanations. Additionally, we avoid discretizing the neural representation as we leverage neural predicates, essential for neatly integrating probability and logic. Lastly, our aim here is different as the paradigm is designed to scale probabilistic NeSy and to quantify the quality-speed tradeoff.
 
\textbf{Neural inference strategies}. The work from~\cite{cornelio2023learning} proposes a neurosymbolic pipeline consisting of a symbolic engine and three neural modules. Specifically, a perception network mapping the images to their symbolic representations, a neural solver attempting to correct the symbolic representation, and a mask predictor to identify possible mistakes done by the neural solver. The symbolic solver then corrects the mistakes on the neural predictions. The neural modules are pre-trained under supervision and fine-tuned using reinforcement learning. In contrast, our simplified framework is based on a scalable logic engine and single neural component. Learning is performed by directly supervising the neural component using the sampled explanations using variational inference strategies. The work in~\cite{krieken2023nesi} introduces two neural modules for neural-symbolic learning: a perception component mapping input data to the probabilities of facts, and a neural reasoner component mapping probabilities to the query. Learning involves training the neural reasoner to mimic synthetic input/output pairs obtained through logical inference, and then training the perception component in a supervised manner using the frozen neural reasoner. In contrast, our work only requires a perception component and utilizes a sampling algorithm that guarantees logically consistent solutions while ensuring diversity. The Neural Theorem Prover~\citep{minervini2020differentiable,minervini2020learning} introduces a continuous and differentiable relaxation of the backward-chaining logic reasoning algorithm. In contrast, our approach does not rely on relaxations or differentiability through the logic program. Other neural sampling strategies, such as GFlowNets~\citep{bengio2021gflownet,bengio2021flow,zhang2022generative}, treat sampling as a sequential decision-making process and learn a policy based on a reward function. However, sampling with hard constraints and exploring solution modes from logical programs remains a challenging problem~\citep{ermon2012uniform,sansone2022lsb}.

\textbf{Logical inference/sampling strategies}. Probabilistic logical inference can be performed exactly by transforming the logical program into a probabilistic circuit (PC) through knowledge compilation~\citep{darwiche2002knowledge}. This allows for efficient evaluation in polynomial time~\citep{choi2020probabilistic,xusemantic,ahmed2022semantic,ahmed2023semantic}. Alternatively, approximate strategies can be used to avoid the computational burden of knowledge compilation, but they may result in biased learning~\citep{manhaeve2021approximate,huangscallop,skryagin2022neural} and in lacking guarantees on the uniformity/quality of the sampled solutions~\citep{jerrum1986random,bellare2000uniform}. The assumption of uniform distribution of worlds is often made when sampling solutions from a CNF formula, and various uniform samplers have been proposed with theoretical guarantees on query complexity and uniformity~\cite{meel2022counting}. State-of-the-art samplers based on hashing-based methods and SAT solvers achieve approximate uniformity by partitioning the solution space into smaller regions~\citep{een2003extensible,soos2009extending,moura2008z3,soos2020tinted,soos2009extending,yang2021engineering}. In contrast, our algorithm focuses on a stronger criterion than uniformity, namely diversity.

\section{Discussion and Conclusion}

The EXAL method is designed as a scalable solution to learning for neural probabilistic logic. The key idea is to use explanation sampling to propagate the learning signal through the symbolic component. This allows the neural component to be trained fast compared to exact approaches. We provide theoretical guarantees on the approximation error and provide practical methods to reduce this error, in particular by encouraging diversity. Experimentally, EXAL is shown to be competitive with state-of-the-art NeSy methods on larger problems in terms of accuracy while significantly outperforming them in terms of speed.

Despite these benefits, sampling explanations is NP-hard and can be slow for problems with a large explanation space. Exploring the entire space can be expensive, but is often unnecessary in practice. A good training signal can be obtained from a sampled subset of explanations. For example, in the Warcraft pathfinding problem it is unlikely to sample the true grid, but sampling similar grids suffices to properly train the network. Whether or not it is necessary to explore the entire explanation space is of course problem dependent. In the worst case where the explanation space is large and only one explanation can provide a good training signal, EXAL will also perform poorly. Furthermore, if satisfiability is difficult to check, sampling explanations for a formula is also difficult, as the existence of an explanation implies satisfiability. NeSy methods, including EXAL, can also suffer from reasoning shortcuts, an issue identified experimentally in~\cite{manhaeve2018deep}. It will be interesting to address this general NeSy issue in future work. Furthermore, this work focuses on scaling learning with a fixed logical formula, but scalability for structure learning is also an open problem. Lastly, EXAL supports continuous inputs $x$, but does not support continuous variables in the symbolic representation $f$. Extending EXAL with continuous $f$ is another line of research that would lead into the field of satisfiability modulo theories and weighted model integration~\cite{morettin2017}.

Although this work focused on NeSy-WMC for the symbolic component, EXAL can be used with any symbolic component for which a suitable EXPLAIN algorithm can be devised. This makes the method flexible to be applied to other use cases in the future.

\begin{ack}
This work was supported by the KU Leuven Research Fund (C14/18/062), the Flemish Government under the ``Onderzoeksprogramma Artifici\"ele Intelligentie (AI) Vlaanderen'' programme and the EU H2020 ICT48 project “TAILOR” under contract \#952215. L. D. R. receives funding from the FWO project ``Neural Probabilistic Logic Programming'' (reference G097720N). E. S. receives funding from the Horizon Europe research and innovation programme (MSCA-GF grant agreement n° 101149800, DISCWORLD).
\end{ack}

\bibliography{ref}

\newpage

\section*{Appendix: Proof of Loss Bounds}

We provide the full proof of Eq.~\ref{eq:bound} here:
\begin{align}
    -\log P(\mathcal{D}) & = -\log \left( \prod_{i} P(\phi_{i} \mid x_{i}) \right) \nonumber \\
    & = -\sum_{i} \log P(\phi_{i} \mid x_{i}) \nonumber \\
    & = -\sum_{i} \log \sum_{f} P(\phi_{i} \mid f) P(f \mid x_{i}) \nonumber \\
    & = -\sum_{i} \log \sum_{f \in \Phi_{i}} P(f \mid x_{i}) \nonumber \\
    & = -\sum_{i} \log \sum_{f \in \Phi_{i}} Q_{i}(f) \frac{P(f \mid x_{i})}{Q_{i}(f)} \nonumber \\
    & \leq -\sum_{i} \sum_{f \in \Phi_{i}} Q_{i}(f) \log \frac{P(f \mid x_{i})}{Q_{i}(f)} \nonumber \\
    & = \sum_{i} \text{KL}(Q_{i} | P_{i}) = L \quad .
\end{align}
Recall that $P(\phi_{i} \mid f)$ is an indicator that returns $1$ if and only if $f$ satisfies $\phi_{i}$, limiting the sum to only explanations $f \in \Phi_{i}$. The bound is obtained using Jensen's inequality, which states that $\log(\mathbb{E}_{Q}[X]) \geq \mathbb{E}_{Q}[\log(X)]$. In this case $X = P(f \mid x_{i}) / Q_{i}(f)$.

\section*{Appendix: Experimental Details}

\subsection*{Diversity}

The diversity experiment has been performed on 9 formulas, of which 4 are shown in Figure~\ref{fig:diversity}. These are 3 branch, 3 bottom-up and 3 split formulas. Branch formulas contain implications of the form $f_{0} \Rightarrow f_{1} \lor ... \lor f_{b}$ with $b$ the branching factor. It also contains one clause with $b$ variables in disjunction. The branching factors are 3, 3 and 10 and the formulas have depth of 3, 5 and 3 respectively. The bottom-up formulas have as parameters the fraction of starting variables, which is always set to 0.5, the number of inferred variables, respectively 20, 60 and 60, and the in-degree, respectively 3, 3 and 4. These formulas recursively define new variables in terms of previously existing variables, e.g. $f_{4} \Leftrightarrow f_{1} \land (f_{2} \lor f_{3})$. The split formulas alternate between layers of conjunctions and disjunctions. All 3 programs have a depth of 4 and a conjunction size of 3 whereas the disjunction size varies from 2 to 4 inclusive.

\subsection*{Convergence of Bounds}

To observe the convergence of bounds, 3 formulas have been created, each with 24 variables. The formulas are created so that exactly half of the assignments are explanations. The probabilities of the variables are then varied in order to set the probability of the query to 0.3, 0.8 or 0.5, of which the first two are shown in Figure~\ref{fig:bounds}. For the diverse sampling algorithm we have used EXPLAIN with $\theta = 3$.

\subsection*{MNIST Addition}

\paragraph{Dataset.}
Generating the data for the MNIST addition experiment~\cite{manhaeve2018deep} on two sequences of $N$ digits is a straightforward process. It involves randomly selecting $2 N$ images from the MNIST dataset and concatenating them to create two distinct sequences, each with a length of $N$. To supervise these sequences, we easily obtain the desired values by multiplying the labels of the selected MNIST images by the appropriate power of $10$. We then sum the resulting sequence of values for each number and further sum the two resulting numbers. It is important to note that each MNIST image is only allowed to appear once in the sequences. Hence, the dataset consists of $\lfloor 60000 / 2 N \rfloor$ sequences available for learning. The test set follows a similar procedure, using the test set partition of the MNIST dataset.

\paragraph{Modelling.}
In this experiment, a traditional LeNet~\cite{lecun1998gradient} neural network is utilized. The network architecture consists of two convolutional layers with $6$ and $16$ filters of size $5$, employing ReLU activations. These layers are followed by a flattening operation. Subsequently, three dense layers with sizes of $120$, $84$, and $10$ are employed. The first two dense layers also utilize ReLU activations, while the final layer applies a softmax activation. The network outputs the probabilities indicating the likelihood that each image in the two sequences corresponds to a specific digit.

\paragraph{Hyperparameters.}
For this experiment, we adopted the standard Adam optimizer with a learning rate of $\eta = 10^{-3}$, known for its reliable performance. Other critical hyperparameters include the number of samples drawn by EXAL and the number of epochs for training. 
In all cases, $600$ samples were used, the same number as A-NeSI.
The maximum number of epochs available for training was set to $10$, $20$ and $100$ for $N = 2$, $N = 4$ and $N = 15$, respectively.
Reproduction of the A-NeSI results for MNIST was done using the optimised hyperparameters as reported by~\citet{krieken2023nesi}.

\subsection*{Warcraft Pathfinding}

\paragraph{Hyperparameters.}
For the Warcraft pathfinding experiment we used a batch size of $100$, a learning rate of $0.0001$ and trained for $10000$ iterations. In every iteration we took a grid, used a burn in period of $100$ tile samples and then sampled $300$ tiles, which were actually used for training. Evaluation was done on $100$ maps and the results were averaged over $5$ reruns of this entire procedure.

\paragraph{Logic formula.} The logic formula for the symbolic computation in the Warcraft pathfinding problem is obtained by converting the below logic program into a formula when querying for the atom $formula$:
\begin{align*}
    & path([target], 0). \\
    & path([N1, N2 \mid P], C) \leftarrow edge(N1, N2, C1), \\
    & \hspace{8mm} path([N2 \mid P], C2), C \text{ is } C1 + C2. \\
    & shortest([source \mid P]) \leftarrow path([source \mid P], C), \\
    & \hspace{8mm} \text{not}( path([source \mid AP], AC), AC < C ). \\
    & formula \leftarrow shortest(`\text{path given by label}'). 
\end{align*}

\end{document}